\documentclass{article}
\usepackage{spconf,amsmath,graphicx, array, cite}

\usepackage[pagebackref=true,breaklinks=true,colorlinks,bookmarks=false]{hyperref}

\usepackage{caption}

\usepackage{xspace}

\usepackage{xcolor}
\def\dark_pdf{\pagecolor[rgb]{0.1,0.1,0.1}\color[rgb]{1.0,1.0,0.0}}

\usepackage{multirow}
\usepackage{graphicx}
\usepackage{booktabs}

\usepackage{amssymb}
\usepackage{amsmath}

\usepackage{nccmath}

\usepackage{pifont}

\usepackage{mathtools}

\usepackage{tcolorbox}

\newcommand{\figref}[1]{Figure~\ref{#1}}
\newcommand{\tabref}[1]{Table~\ref{#1}}

\newcommand{\secref}[1]{Section~\ref{#1}}

\newcommand{\veccs}[1]{\ensuremath{\mathbf{#1}}}

\newcommand{\mset}[1]{\ensuremath{\mathcal{#1}}}
\newcommand{\mtensor}[1]{\ensuremath{\mathbf{#1}}}

\newcommand{\Mod}[1]{\ (\mathrm{mod}\ #1)}

\makeatletter
\DeclareRobustCommand\onedot{\futurelet\@let@token\@onedot}
\def\@onedot{\ifx\@let@token.\else.\null\fi\xspace}                
\def\eg{\emph{e.g}\onedot}             
            
\def\ie{\emph{i.e}\onedot}             
            
\def\etal{\emph{et al}\@onedot}          
\makeatother

\def\JandF{$\mathcal{J}\mathcal{\&}\mathcal{F}$\xspace}

\def\AUC{AUC-}
\def\JFinter{R-\AUC\JandF\xspace}
\def\JFinterx{R-\AUC\JandF}
\def\JF@s{$\mathcal{J}\mathcal{\&}\mathcal{F}@60s$\xspace}
\def\J{$\mathcal{J}$\xspace}
\def\F{$\mathcal{F}$\xspace}

\def\S2M{\href{https://github.com/hkchengrex/MiVOS}{S2M}~\cite{Cheng_CVPR_2021}}

\def\DeepLabv3+{DeepLabv3+~\cite{Chen2018}}
\def\ResNet50{ResNet50~\cite{He2016}}

\def\BL30K{{\href{https://github.com/hkchengrex/MiVOS}{BL30K}}~\cite{Cheng_CVPR_2021}}

\def\Pascal_VOC{PASCAL VOC~\cite{Everingham2011}}

\def\sota{state-of-the-art\xspace}

\title{Revisiting Click-based Interactive Video Object Segmentation}
\name{{\em St\'ephane Vujasinovi\'c$^{1}$, Sebastian Bullinger$^{1}$, Stefan Becker$^{1}$}\\{\em Norbert Scherer-Negenborn$^{1}$, Michael Arens$^{1}$ and Rainer Stiefelhagen$^{2}$}}
\address{
$^{1}$Fraunhofer Institute of Optronics, System Technologies and Image Exploitation (IOSB)$^\star$\thanks{$^\star$Fraunhofer IOSB is a member of the Fraunhofer Center for Machine Learning.}
\\
$^{2}$Karlsruhe Institute of Technology (KIT)
}

\begin{document}
	
\ninept

\maketitle
\begin{abstract}
	While current methods for \emph{interactive Video Object Segmentation}~(iVOS) rely on scribble-based interactions to generate precise object masks, we propose a \emph{Click-based interactive Video Object Segmentation} (CiVOS) framework to simplify the required user workload as much as possible. CiVOS builds on de-coupled modules reflecting user interaction and mask propagation. The interaction module converts click-based interactions into an object mask, which is then inferred to the remaining frames by the propagation module. Additional user interactions allow for a refinement of the object mask. The approach is extensively evaluated on the popular interactive~DAVIS dataset, but with an inevitable adaptation of scribble-based interactions with click-based counterparts. We consider several strategies for generating clicks during our evaluation to reflect various user inputs and adjust the DAVIS performance metric to perform a hardware-independent comparison. The presented CiVOS pipeline achieves competitive results, although requiring a lower user workload.
\end{abstract}

%

\section{Introduction}
\label{sec:intro}
\subsection{Motivation}
\emph{Video Object Segmentation} (VOS) is a fundamental and challenging problem in computer vision, where the objective is to segment an object of interest in a video sequence. An abundance of methods have been proposed over the years, which can be classified according to~\cite{Wang2021} into three categories: \emph{automatic}~(aVOS), \emph{semi-automatic}~(sVOS) and \emph{interactive}~(iVOS) methods. In the context of aVOS, the goal is to automatically perform VOS without human intervention, limiting the usage to only known objects. In sVOS, human intervention is allowed but limited, as only a single frame is annotated with the full mask of the object of interest before performing VOS.

In comparison to aVOS and sVOS approaches, iVOS methods (which are the focus of this paper) are concerned with the integration of continuous user interactions during the VOS process, to ensure high reliability while segmenting arbitrary objects.
Recent methods incorporate user annotations (\ie, scribbles) in a multi-round setting through the following design: prediction of segmentation masks based on user annotations and temporal propagation of the predicted masks to the remaining video frames~\cite{benard2017interactive, Oh_CVPR_2019,Heo_ECCV_2020, Miao_CVPR_2020,Cheng_CVPR_2021}. 
Direct applications can be found in the entertainment and dataset collection industry, where annotating a huge amount of data on pixel-level is tedious and time-consuming.

The primary goal of iVOS approaches is to segment an arbitrary object instance with minimal user effort. To this end, using a click-based approach (instead of scribbles) allows the user to designate an object of interest rapidly~\cite{bearman2016s, benenson2019large} and more intuitively~\cite{bearman2016s}, while demanding a minimal annotation effort.
As highlighted by Bearman~\etal~\cite{bearman2016s}, a user needs approximately $11$ seconds to annotate an object instance with a scribble (depending on the level of details~\cite{Caelles_arXiv_2018}), while a click costs the user between $1$ to $3$ seconds~\cite{bearman2016s, benenson2019large}.
Here, \emph{interactive Image Object Segmentation} (iIOS) approaches are presumably helpful, as recent methods have demonstrated impressive results leveraging click-based interactions to predict masks for arbitrary object instances in single images~\cite{xu2016deep,liew2017regional,mahadevan2018iteratively,li2018interactive,song2018seednet,Jang2019,Sofiiuk2020,Lin_CVPR_2020,kontogianni2020continuous,Sofiiuk2021}.

\subsection{Contribution}
In this work, we introduce a novel \emph{Click-based interactive Video Object Segmentation} (CiVOS) pipeline that achieves competitive results to scribble-based approaches, while significantly reducing the workload of user interactions.
To reflect the conditions of real-world~iVOS scenarios, we consider several, carefully selected strategies for generating click-based user interactions. We propose a hardware-independent adaptation (\mbox{\JFinterx}) of the current standard metric (\mbox{\AUC\JandF}) to evaluate~iVOS approaches. This allows us to overcome the limitations of evaluations in previous works and report reproducible results. By performing an exhaustive quantitative evaluation on the interactive track of the popular DAVIS~\cite{Perazzi_CVPR_2016, Pont-Tuset_arXiv_2017, Caelles_arXiv_2018, Caelles_arXiv_2019} benchmark, we demonstrate the effectiveness of the proposed pipeline~(\mbox{\JFinterx$=0.76$}, \mbox{\AUC\JandF~$=0.83$}).

We make our source code for CiVOS and the adjusted evaluation framework publicly available to facilitate future research\footnote{Our source code is available at \url{https://github.com/Vujas-Eteph/CiVOS}.}.

\section{Relate Work}
\label{sec:relatework}
\subsection{Interactive Video Object Segmentation}
Early methods for addressing iVOS use hand crafted features (\eg, click-based interactions~\cite{Wang2014TouchCutFI, jain2016click}) but they require a large number of user interactions, making them impractical at a large scale~\cite{Wang2021}.
With the release of the DAVIS interactive benchmark~\cite{Caelles_arXiv_2018}, a standardization for the evaluation of iVOS methods has been introduced, where a key aspect is to identify methods that can be used in real-world applications. 
A majority of recent methods focus on scribble-based interactions as it is the default user annotation provided by the DAVIS benchmark, with the exception of two studies allowing for click-based interactions~\cite{chen2018blazingly, Cheng_CVPR_2021}.

The current practice in state-of-the-art methods is to tackle the problem following the \emph{Interaction-Propagation} design introduced by Benard \etal in~\cite{benard2017interactive}. 
The authors combine two separate networks: an interactive segmentation network based on~\cite{xu2016deep} that predicts and refines segmentation masks based on the interactions of the user as well as a sVOS network~\cite{caelles2017one} which propagates the predicted masks derived from the interaction module to the remaining video frames. 
Oh \etal~\cite{Oh_CVPR_2019} connect the interaction and propagation module with an additional component (\ie, feature aggregation), which requires a joint training of all modules.
Heo \etal~\cite{Heo_ECCV_2020} use a global and local transfer mechanism during propagation for conveying segmentation results specifically to adjacent and distant frames. 
A more efficient approach is explored by Miao \etal~\cite{Miao_CVPR_2020} by using a single backbone network for the interaction and the propagation modules.
A modular architecture on the popular \emph{Interaction-Propagation} design is proposed by Cheng \etal~\cite{Cheng_CVPR_2021} and where a third neural network (\emph{Difference-Aware Fusion}) is added to capture and integrate masks differences between each round.

An alternative to the \emph{Interaction-Propagation} design was explored by Chen~\etal~\cite{chen2018blazingly}, who approached the problem by describing it as a pixel-wise classification task (\ie, k-nearest neighbors in an embedding space) using convolutional neural networks. Recently Chen~\etal~\cite{chen2020scribblebox} presented an interactive annotation tool, where the user annotates objects using tracked boxes and scribbles.

More recent works focus on the applicability of iVOS methods for real-world scenarios by designing networks that automatically determine suitable frames for the user to annotate. 
For instance, Heo \etal~\cite{heo2021guided} use a reliability score for each propagated mask, to select candidate frames for the user. 
Yin \etal~\cite{yin2021learning} question the current iVOS paradigm, where the worst segmented fame is annotated. Instead, they assess the potential of each frame to improve the global segmentation results through a reinforcement learning framework that can be added to existing iVOS methods.

\subsection{Interactive Image Object Segmentation}
The goal in iIOS is to predict an accurate mask of an arbitrary object instance using minimal user annotations (\eg, bounding points, scribbles, clicks).
Xu \etal~\cite{xu2016deep} introduce deep learning in the context of iIOS for the extraction of salient features. By relying on click-based interactions the authors can simulate user inputs and generate a large number of training samples.
Liew \etal~\cite{liew2017regional} extend the previous work by exploiting the context of local regions around user interactions.
Mahadevan \etal~\cite{mahadevan2018iteratively} introduce iterative training for interactive segmentation networks.
Li \etal~\cite{li2018interactive} consider the multi-modal nature of iIOS and propose a network to infer multiple masks in order to cover all plausible solutions as well as a second network to select a result from the set of potential object masks. 
Song \etal~\cite{song2018seednet} use a \emph{Markov Decision Process} trained with reinforcement learning to automatically generate interactions based on the initial interactions of the user.
Jang \etal~\cite{Jang2019} propose the \emph{Back-propagation Refinement Scheme} (BRS) to constrain their network to predict correct labels at user-specified locations, by optimizing the interaction maps (\ie, input of the network) through forward and backward passes. 
Sofiuuk \etal~\cite{Sofiiuk2020} elaborate the idea further and introduce \emph{feature}-BRS (\emph{f}-BRS). Instead of optimizing the interaction maps, \emph{f}-BRS optimizes auxiliary parameters in the last layers of the network. This reduces the computational time during the forward and backward passes. 
Lin \etal~\cite{Lin_CVPR_2020} emphasize the role of the first interaction, by introducing a first click attention module, which leverages the first click as guidance for the following interactions. 
Kontogianni \etal~\cite{kontogianni2020continuous} treat user corrections as training samples to update the network parameters, which also enables cross-domain adaptation. 
Sofiiuk \etal~\cite{Sofiiuk2021} expand the iterative training strategy proposed in~\cite{mahadevan2018iteratively} by introducing a new loss function and an encoding layer that allows integrating additional external information without affecting the pre-trained weights of the backbone encoder.

\section{Click-based Interactive Video Object Segmentation}
\label{sec:method}

\def\Mask{M}
\def\Img{I}
\def\TargetImg{\mtensor{\Img}_{i(r)\pm s}}
\def\TargetMask{\mtensor{\Mask}_{i(r)\pm s}}

\begin{figure*}
	\centering
	\includegraphics[width=1\textwidth]{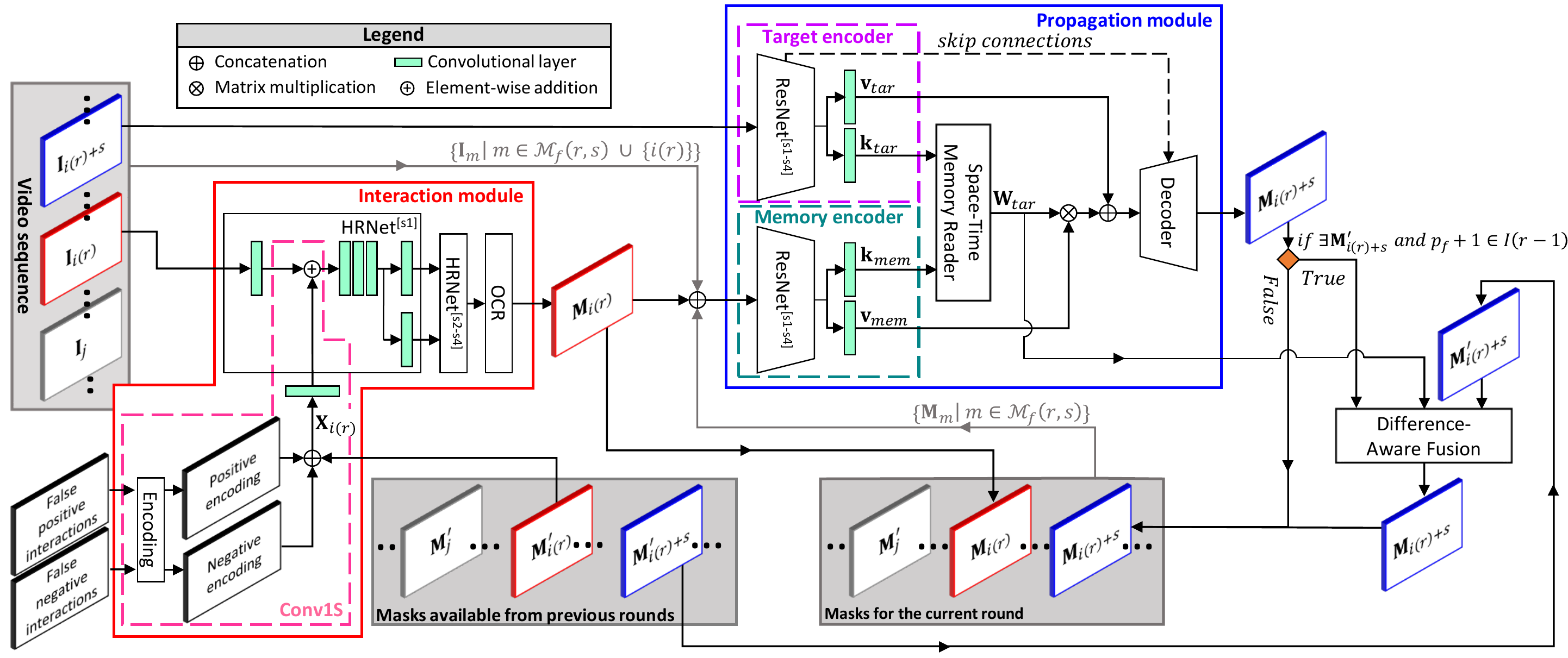}
	\caption{An illustration of our CiVOS architecture. To maintain clarity, we illustrate the process for the forward propagation of round~$r$ at step~$s$. 
	The variables~$\mtensor{\Img}_{i(r)}$ and~$\mtensor{\Mask}_{i(r)}$ denote the image and the derived object mask at~$i(r)$, where~$i(r)$ represents the index of the user interaction in round $r$. 
	The mask~$\mtensor{\Mask}_{i(r)}$ is iteratively propagated to its adjacent frames $i(r)+s$ resulting in~$\mtensor{\Mask}_{i(r)+s}$. 
	Here,~$\{j \mid j \notin \{i(r)\} \cup \mset{P}(r)\}$ denotes the index of a frame that is neither an annotated frame nor a frame used for propagation.}
	\label{fig architectur}
\end{figure*}

VOS can be defined as a binary classification problem that aims to distinguish pixels of a specific object from other (background) structures. Thus, incorrectly classified pixels can be regarded as either false positives or as false negatives. To rectify misclassified pixel regions, the user may indicate false negative and false positive regions with corresponding interactions. 
As illustrated in~\figref{fig architectur}, we build on the modular concept introduced by~MiVOS~\cite{Cheng_CVPR_2021} for the proposed CiVOS framework, which comprises two fundamental deep learning architectures, an interaction and a propagation module. 

Given a click-based user input (\ie, false positive and false negative interactions) for a frame $\mtensor{\Img}_{i(r)}$, the interaction module predicts a corresponding segmentation mask $\mtensor{\Mask}_{i(r)}$, where $r$ denotes the current round and $i(r)$ the index of the frame annotated in round $r$. Subsequently, the propagation module propagates the mask $\mtensor{\Mask}_{i(r)}$ in both directions (backward and forward) starting from the annotated frame~$\mtensor{\Img}_{i(r)}$. For each direction, the propagation procedure is applied on all images located between $i(r)$ and the closest annotated frame of a previous interaction round. If no user annotation exists, the masks are propagated to the start or the end of the video sequence, respectively.
	
More formally, let $\mset{I}(r) = \{ i(r') \mid r' \in \{1, \dots, r\} \}$ denote the set of all frame indices with user annotations of preceding rounds. 
Further, let $p_b$ be the last frame index in round $r$ used for backward propagation, \ie $p_b = \max(\{ i^{\prime} + 1 \mid i^{\prime} \in \mset{I}(r-1) \land i^{\prime} < i(r) \} \cup \{ j_0 \})$ with $j_0$ representing the index of the first frame in the video sequence. Similarly, we define the last frame index of the forward propagation $p_f$ according to $p_f = \min(\{ i^{\prime} - 1 \mid i^{\prime} \in \mset{I}(r-1) \land i^{\prime} > i(r) \} \cup \{ j_n \})$. 
Here, $j_n$ denotes the index of the last frame in the video sequence. With $\mset{P}_b(r)$ and $\mset{P}_f(r)$, the frame indices for propagation in round~$r$ defined by $\{p_{b}, \dots, i(r)-1\} \cup \{i(r)+1, \dots, p_{f}\} \coloneqq \mset{P}_{b}(r) \cup \mset{P}_{f}(r) \coloneqq \mset{P}(r)$.

In addition, we use~$\mtensor{M}^{\prime}_{j}$ to denote the most recent mask inferred for frame~$\mtensor{\Img}_{j}$ from an interaction, propagation, or fusion step of a previous round, with~$j \in \{j_0, \ldots, j_n\}$.

\subsection{Interaction Module}
We consider RITM~\cite{Sofiiuk2021} as the interaction module since it is (presumably) the current \sota method for tackling iIOS problems using click-based interactions. HRNet+OCR~\cite{HRNet,OCR} serve as the backbone for the segmentation network. 

As illustrated in~\figref{fig architectur}, the inputs for the interaction module are: the current frame used for interactions $\mtensor{\Img}_{i(r)}$, two encoded interaction maps~\cite{benenson2019large,Sofiiuk2021} that represent incorrectly segmented or missing mask areas and the latest available mask~$\mtensor{\Mask}^{\prime}_{i(r)}$.
We use a convolutional block (\ie, \emph{Conv1S}~\cite{Sofiiuk2021}) as illustrated in~\figref{fig architectur} to take additional external inputs into account without affecting the pre-trained weights of the backbone segmentation network (\ie, HRNet~\cite{HRNet}). 
Here, the mask~$\mtensor{\Mask}^{\prime}_{i(r)}$ is concatenated channel-wise with the interaction maps into~$\mtensor{X}_{i(r)}$, before passing through a convolutional layer.
The output of the convolutional layer~(\ie,~\emph{Conv1S}) is a tensor with the same dimensions as the tensor given by the first layer of HRNet, such that an element-wise summation can be applied along the channel axis.
As in \emph{f}-BRS~\cite{Sofiiuk2020} the predicted mask is optimized to predict the user-specified labels at the locations where the interactions occurred.

\subsection{Propagation Module}
We consider the propagation branch of MiVOS~\cite{Cheng_CVPR_2021}, inspired by~\cite{Oh_CVPR_2019} as the mask propagation component, which relies on a two-encoder- and one-decoder-structure as illustrated in~\figref{fig architectur}.

\begin{figure}
	\centering
	\includegraphics[width=0.48\textwidth]{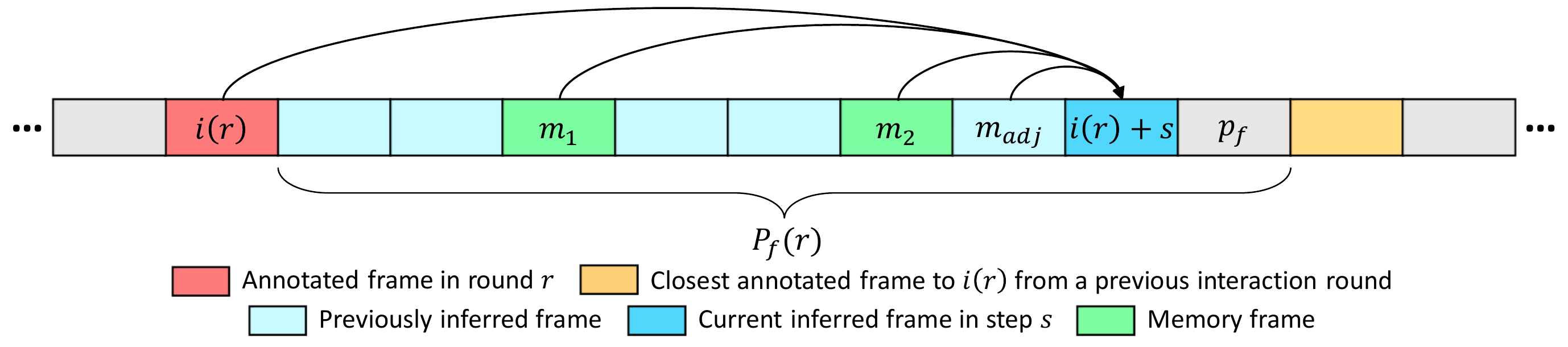}
	\caption{Illustration of the propagation scheme with~$d=3$. For simplicity, we only display the forward propagation and omit the fusion step. We denote with $m_{adj}$ the index of the frame inferred in the previous step $s-1$ such that $m_{adj}=i(r)+(s-1)$.}
	\label{fig 2}
\end{figure}

The backbone feature extractor for both encoders (\ie, the target encoder and the memory encoder) use ResNet$50$~\cite{He2016} up to and including the fourth stage (\ie, resnet layer \emph{conv$4$\_$5$}), which computes $1024$ different feature channels.
The propagation is a step-wise process, where only one frame is segmented at a time as illustrated in~\figref{fig 2}. In each step $s\in \mathbb{N}^{\star}$, the frame $\mtensor{\Img}_{i(r)\pm s}$ is fed into the target encoder (the addition or subtraction depends on the direction of the propagation). 
In contrast to the target encoder, the memory encoder takes into account several frames (with their respective masks) denoted as memory frames. 
The corresponding frame indices $\mset{M}(r,s)$ depend on the index of the user interaction $i(r)$ and the current step. 
Concretely, the indices of the memory frames (and their respective masks) used as input for the memory encoder in round $r$ and step $s$ are defined by 
$\mset{M}(r,s) = \{ i(r)\} \cup \mset{M}_{b}(r,s) \cup \mset{M}_{f}(r,s)$ with 
$\mset{M}_{b}(r,s) = \{m \in \mset{P}_{b}(r) \mid i(r) > m > i(r)-s \land m \equiv i(r)\ (\textrm{mod}\ d)\} \cup \{i(r)-(s-1)\} \cap \{{P}_{b}(r)\}$ and 
$\mset{M}_{f}(r,s) = \{m \in \mset{P}_{f}(r) \mid i(r) < m < i(r)+s \land m \equiv i(r)\ (\textrm{mod}\ d)\} \cup \{i(r)+(s-1)\} \cap \{{P}_{f}(r)\}$.
We denote with \mbox{$m \equiv i(r) \Mod{d}$} the condition that every memory frame index~$m$ should be distant from the user interaction~$i(r)$ by a multiple of~$d$, where $d \in \mathbb{N}^{\star}$. 

Two separate convolution layers are added at the end of each encoder, as shown in~\figref{fig architectur}, to extract key-values features \mbox{$\{  \{\veccs{k},\veccs{v}\} \mid \veccs{k} \in \mathbb{R}^{c_{k}\times hw}, \veccs{v} \in \mathbb{R}^{c_{v}\times hw} \}$}, in order to make use of an attention-based operation. 
Here, $c$ designates the channel dimensions and $h \times w$ the feature map resolution. 
Since the memory encoder accepts $|\mset{M}(r,s)|$ memory frames (with their respective masks), the dimensions for the memory key-values features are~$\veccs{k}_{mem} \in \mathbb{R}^{c_{k}\times |\mset{M}(r,s)|hw}$~and~$\veccs{v}_{mem} \in \mathbb{R}^{c_{v}\times |\mset{M}(r,s)|hw}$. 

A similarity matching between key features, $\veccs{k}_{tar}$ and $\veccs{k}_{mem}$ is performed in the \emph{Space-Time-Memory Reader} following~\cite{Cheng_CVPR_2021}, to infer an attentional weight matrix $\mtensor{W}_{tar}\in\mathbb{R}^{|\mset{M}(r,s)|hw\times hw}$, indicating which features from $\veccs{k}_{tar}$ show the highest similarity to $\veccs{k}_{mem}$. 
The attentional weight matrix $\mtensor{W}_{tar}$ allows us to weight the memory values $\veccs{v}_{mem}$ only at relevant feature positions. With~$\veccs{v}_{mem}$ encoding which features belong to the foreground or background. 
The weighted memory values $\veccs{v}_{mem}$ are then concatenated with the target values $\veccs{v}_{tar}$, which encode appearance information of the frame~$\TargetImg$.
The resulting concatenation is fed to the decoder in order to propagate the mask of the object of interest to $\TargetMask$.

The decoder is based on~\cite{Oh_CVPR_2019} and employs several refinement modules~\cite{oh2018fast} as building blocks. At each stage, a refinement module takes the up-sampled feature output from the previous refinement module along with the feature maps of the corresponding target encoder via skip connections.
 
Since the propagation process is independent of the masks inferred in previous rounds $\mtensor{M}^{\prime}$, the information contained in those masks and thus the intent of the previous user interactions may be lost.
To prevent this, a learnable fusion component (\emph{Difference-Aware Fusion}) based on~\cite{Cheng_CVPR_2021} as shown in~\figref{fig architectur}, infers a new mask $\mtensor{\Mask}_{i(r)\pm s}$. 
A fusion between $\mtensor{\Mask}_{i(r)\pm s}$ and $\mtensor{\Mask}^{\prime}_{i(r)\pm s}$ is performed during the propagation in the backward direction if $p_{b} -1 \in \mset{I}(r-1)$ and vice versa in the forward direction if~$p_{f} +1 \in \mset{I}(r-1)$.

\section{Dataset and Experiments}
\label{sec:dataset}
\subsection{DAVIS Interactive Benchmark}
To evaluate CiVOS, we utilize the popular interactive 2017 DAVIS benchmark~\cite{Perazzi_CVPR_2016, Pont-Tuset_arXiv_2017, Caelles_arXiv_2018,Caelles_arXiv_2019} containing high-quality segmentation annotations~\cite{Pont-Tuset_arXiv_2017}. 
Here, a robot agent simulates scribble-based user interactions. 
Since the process is round-based, the robot annotates at the beginning of each round~$r$ a single frame with one or several scribbles. 
At the end of the round, the evaluation algorithm compares the predicted results against the ground-truth annotations in order to determine the frame with the worst segmentation and provide corrective scribbles. 

The benchmark uses a maximum of $8$ interactions rounds while evaluating the submitted iVOS approaches with a time budget (which is set to $30$ seconds per round for each object in the sequence). 
We~also use these values in our evaluations reported in~\tabref{tab_CiVOS_vs_SOTA} and~\tabref{tab_CiVOS_clicks}.

The DAVIS benchmark uses the following metrics~\cite{Perazzi_CVPR_2016}: a region similarity~\J~(\ie, the Jaccard index) and a contour accuracy~\F. 
To express both metrics with a single number the average of~\J and~\F is summarized as~\JandF. 
To aggregate the results of several successive interactions the DAVIS benchmark uses the area under the curve of~\JandF~(denoted as~\AUC\JandF). 
The required time to predict the masks is used as~$x$-axis. 
The benchmark further introduces the~\JF@s metric to denote the obtained~\JandF score at $60$~seconds.

\subsection{Metric Adaptation and Click Simulation}
\label{point extraction}
The \AUC\JandF, as well as the \JF@s metric, incorporate the processing time to promote fast methods suitable for real-world applications. 
Thus, the evaluation scores are hardware-dependent and not reproducible on different systems. 
The metrics favor high-end machines as they are able to achieve high \JandF scores faster in the allowed time budget. 
To enable a fair and reproducible comparison across different systems, we propose \JFinter that adapts the \AUC\JandF metric by incorporating the accumulated interaction rounds instead of the processing time.

As we consider a click-based approach to tackle iVOS, we propose three new click-based interaction generation strategies denoted as~$f_{1}$,~$f_{2}$,~$f_{3}$ for the interactive DAVIS framework. The strategies~$f_{1}$ and~$f_{2}$ rely on the generated scribbles of the robot agent to extract click-based inputs. 
In the case of~$f_{1}$, in each round a single click is generated for each object based on all assigned scribbles to the corresponding object. In contrast,~$f_{2}$ computes a click for each scribble. 
We consider the scribbles of the same object as an ensemble of points with their respective~$(x,y)$ coordinates. 
With~$f_{1}$ we infer the average position of all points and with~$f_{2}$ the average positions of the points belonging to the same scribble.
Afterward, the closest scribble coordinate~$(x,y)$ to an inferred average point is used as a click for the user input simulation.
Lastly, we exploit directly the erroneous regions between prediction and ground-truth to emulate novel user interactions~($f_{3}$). 
For this strategy, the simulated clicks are located at the center of the corresponding regions.

\subsection{Quantitative Results}
\label{sec:experiments}
To provide a fair comparison of CiVOS against current \sota methods, we generate click-based interactions from the scribbles provided by the DAVIS robot agent following the~$f_{2}$ strategy presented in~\secref{point extraction}. Since the first interaction round only includes one scribble per object, we also utilize only one click per object regardless of the click generating strategy used~(\ie, $f_{1}$,~$f_{2}$,~$f_{3}$).

We compare CiVOS against \sota methods relying on scribble-based interactions on the interactive 2017 DAVIS benchmark~\cite{Perazzi_CVPR_2016, Pont-Tuset_arXiv_2017, Caelles_arXiv_2018,Caelles_arXiv_2019} in~\tabref{tab_CiVOS_vs_SOTA}.
Specifically, we compare our approach against a MiVOS~\cite{Cheng_CVPR_2021} variation that accepts click-based interactions.
CiVOS achieves competitive results~(\ie,~\mbox{\JFinterx$=0.76$}, \mbox{\AUC\JandF~$=0.83$}) to scribble-based methods despite relying only on click-based interactions.

\tabref{tab_CiVOS_clicks} displays the \JFinterx results of CiVOS on the DAVIS 2017 validation set by taking into account different click-based interaction strategies~(\ie, $f_{1}$,$f_{2}$,$f_{3}$). The number of interactions allowed per object per round depends on the extent of the erroneous regions, \ie, small incorrectly classified areas are not considered during this process.
We are able to boost the performance of~CiVOS~(\ie,~\mbox{\JFinterx$=0.78$}) by extracting directly the central coordinates of the misclassified regions rather than arbitrarily placed points (points generated from scribbles might be located on the object edge or far from the object center) of the erroneous regions. 
Also, we do not observe any improvement of the~\JFinterx score after a third interaction on the same object within the same round of interaction.
Thus, limiting the number of interactions up to a maximum of three interactions per round and per object provides a good trade-off between user workload and mask accuracy.

\begin{table}[h]
	\begin{center}
		\caption{Quantitative evaluation on the interactive DAVIS 2017 validation set.
		We evaluate current \sota scribble-based iVOS methods with the newly presented \JFinterx metric for reproducible comparison and test CiVOS and MiVOS on click-based interactions (using the $f_{2}$ strategy).}
		\resizebox{\columnwidth}{!}{
			\begin{tabular}{lccccccc}
				\hline
				\multirow{2}{*}{Methods} 			& Training 	    & Testing	        & R-\AUC				& \AUC  				& \JandF   			\\
													& Interaction	& Interaction       & \JandF$\uparrow$		& \JandF$\uparrow$  	& $@60s\uparrow$	\\
				\hline\hline
				MANet~\cite{Miao_CVPR_2020}			& Scribble 	    & Scribble 		        & $0.72$ 				& $0.79$ 				& $0.79$ 			\\
				ATNet~\cite{Heo_ECCV_2020}			& Scribble 	    & Scribble			    & $0.75$ 				& $0.80$				& $0.80$ 			\\
				MiVOS~\cite{Cheng_CVPR_2021}		& Scribble 	    & Scribble			    & $\mathbf{0.81}$		& $\mathbf{0.87}$		& $\mathbf{0.88}$	\\
				GIS-RAmap~\cite{heo2021guided}		& Scribble 	    & Scribble     	        & $0.79$ 				& $0.86$ 				& $0.87$ 			\\
				\hline
				MiVOS~\cite{Cheng_CVPR_2021}		& Click			& Click			    & $0.70$				& $0.75$				& $0.75$			\\
				CiVOS(\textbf{Ours})				& Click		   	& Click			    & $\mathbf{0.76}$		& $\mathbf{0.83}$		& $\mathbf{0.84}$	\\
				\hline
			\end{tabular}
		}
		\label{tab_CiVOS_vs_SOTA}
	\end{center}
\end{table}

\begin{table}[h]
	\begin{center}	
		\caption{\JFinterx results on the DAVIS 2017 validation set by using different
			click generating strategies. CiVOS achieves the best result by incorporating at most three clicks for each object per round and by relying on the central coordinates of erroneous regions.}
		\resizebox{\columnwidth}{!}{
			\begin{tabular}{lccccccc}
				\hline
				
				Maximal Number   &\multirow{2}{*}{$1$}	  	&\multirow{2}{*}{$2$}	&\multirow{2}{*}{$3$}	&\multirow{2}{*}{$4$}	&\multirow{2}{*}{$5$}	&\multirow{2}{*}{$6$}	&\multirow{2}{*}{$7$}	\\
				of Clicks  & & & & & & & \\
			
				\hline\hline
				Interaction Strategy~$f_{1}$ 	& $0.69$	& $-$	  				& $-$	  				& $-$					& $-$	  				& $-$	  				& $-$		\\
				Interaction Strategy~$f_{2}$ 	& $0.72$	& $\mathbf{0.76}$  		& $0.76$				& $0.75$				& $0.75$  				& $0.75$				& $0.76$	\\
				Interaction Strategy~$f_{3}$ 	& $0.74$	& $0.77$ 		 		& $\mathbf{0.78}$		& $0.78$  				& $0.78$				& $0.78$  				& $0.78$	\\
				\hline
			\end{tabular}
		}
		\label{tab_CiVOS_clicks}
	\end{center}
\end{table}

\section{Conclusion}
\label{sec:conclusion}
We propose CiVOS, for reconsidering click-based approaches for iVOS, to effectively reduce the workload of the user while maintaining competitive performance against scribble-based approaches, as indicated by the evaluation on the DAVIS validation set. 
We also present an adaptation of the \AUC\JandF metric by using the interaction rounds denoted as \JFinterx, enabling the repeatability of the evaluations across different systems.


\bibliographystyle{IEEEbib}
\bibliography{CiVOS_arXiv}


\end{document}